\documentclass[fleqn,10pt]{wlscirep}
\usepackage[utf8]{inputenc}
\usepackage[T1]{fontenc}
\usepackage{ifthen}
\usepackage{pgfplots}
\usepackage{calc}
\usepackage{booktabs}
\usepackage{gensymb}
\usepackage[position=top, labelformat=simple]{subfig}
\usepackage{tikz}
\usepackage{amsmath,amssymb,amsfonts}
\usepackage{algorithmic}
\usepackage{graphicx}
\usepackage{textcomp}
\usepackage{array}
\usepackage{xcolor}
\usepackage{upgreek}
\usepackage{floatrow}
\usepackage[acronym]{glossaries}
\usepackage{siunitx}
\usepackage{svg}

\hypersetup{allcolors = black}

\usepackage{lineno}

\newcolumntype{C}{>{\centering\arraybackslash} m{6cm} }

\usepackage[acronym]{glossaries}
\makeglossaries
\newacronym{casa}{CASA}{computer-aided sperm analysis}
\newacronym{who}{WHO}{World Health Organization}
\newacronym{cnn}{CNN}{convolutional neural network}
\newacronym{ml}{ML}{machine learning}
\newacronym{mae}{MAE}{mean absolute error}
\newacronym{mse}{MSE}{mean squared error}
\newacronym{rmse}{RMSE}{root mean squared error}
\newacronym{hushem}{HuSHeM}{Human Sperm Head Morphology}
\newacronym{tfr}{TFR}{total fertility rate}
\newacronym{bmi}{BMI}{body mass index}
\newacronym{sma}{SMA}{sperm morphology analysis}
\newacronym{lr}{LR}{linear regression}
\newacronym{rf}{RF}{random forest}
\newacronym{zeror}{ZeroR}{zero regression}
\newacronym{lstm}{LSTM}{Long short-term memory}
\newacronym{jpdaf}{JPDAF}{joint probabilistic data association filter}
\newacronym{gdpr}{GDPR}{General Data Protection Regulation}

\title{Machine Learning-Based Analysis of Sperm Videos and Participant Data for Male Fertility Prediction}

\author[1,2*]{Steven A. Hicks}
\author[3+]{Jorunn M. Andersen}
\author[3+]{Oliwia Witczak}
\author[1,2]{Vajira Thambawita}
\author[1,2]{Paal Halvorsen}
\author[2]{Hugo L. Hammer}
\author[3++]{Trine B. Haugen}
\author[1,4++]{Michael A. Riegler}
\affil[1]{Holistic Systems Department, Simula Metropolitan Center for Digital Engineering, Oslo, Norway}
\affil[2]{Faculty of Technology, Art and Design, OsloMet – Oslo Metropolitan University, Oslo, Norway}
\affil[3]{Faculty of Health Sciences, OsloMet – Oslo Metropolitan University, Oslo, Norway}
\affil[4]{Department of Technology, Kristiania University College, Oslo, Norway}
\affil[*]{contact/corresponding author steven@simula.no}
\affil[+]{contributed equally to the work}
\affil[++]{joint senior authors}


\begin{abstract} 
\textbf{Preprint - Original to appear in Nature Scientific Reports}

Methods for automatic analysis of clinical data are usually targeted towards a specific modality and do not make use of all relevant data available. In the field of male human reproduction, clinical and biological data
are not used to its fullest potential. Manual evaluation of a semen sample using a microscope is time-consuming and requires extensive training. Furthermore, the validity of manual semen analysis has been questioned due to limited reproducibility, and often high inter-personnel variation. The existing computer-aided sperm analyzer systems are not recommended for routine clinical use due to methodological challenges caused by the consistency of the semen sample. Thus, there is a need for an improved methodology.
We use modern and classical machine learning techniques together with a dataset consisting of $85$ videos of human semen samples and related participant data to automatically predict sperm motility. Used techniques include simple linear regression and more sophisticated methods using convolutional neural networks. Our results indicate that sperm motility prediction based on deep learning using sperm motility videos is rapid to perform and consistent. The algorithms performed worse when participant data was added. In conclusion, machine learning-based automatic analysis may become a valuable tool in male infertility investigation and research.
\end{abstract}

\begin{document}

\flushbottom
\maketitle

\thispagestyle{empty}


\section*{Introduction}\label{section:introduction}

Automatic analysis of clinical data may open new avenues in medicine, though often limited to one modality, usually images~\cite{topol2019high}. Recently, however, trends have shifted to include data from other modalities, including sensor data and participant data~\cite{boll2018health, Riegler2016}. Furthermore, advancements in artificial intelligence, specifically deep learning, have shown its potential in becoming an essential tool for health professionals through its promising results on numerous use-cases~\cite{topol2019high, Hannun2019, esteva2017, Pogorelov2017}.



\begin{figure}[t]
    \centering
    \includegraphics[width=0.8\columnwidth]{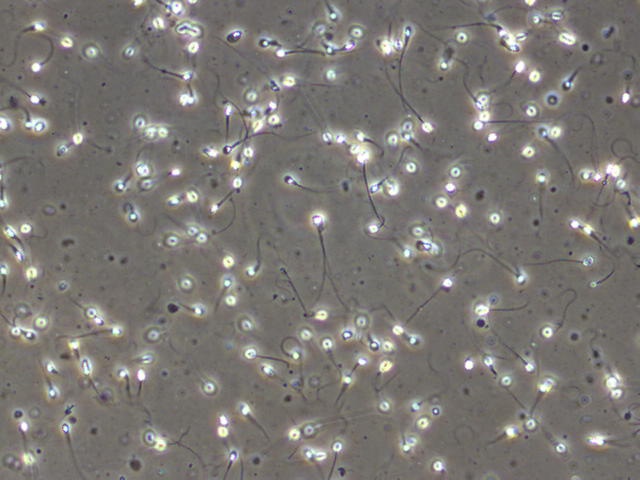}
    \vspace{-5px}  
    \caption{Frame from a microscopic video of a human semen sample showing several spermatozoa (Olympus CX31 phase contrast microscope with heated stage, UEye UI-2210C camera, 400x magnification).}
    \label{fig:sperm}
\end{figure}

Male reproduction is a medical field that is gaining increased attention due to several studies indicating a global decline in semen quality during the last decades~\cite{carlsen1992evidence, levine2017temporal} as well as geographical differences~\cite{jorgensen2002east}. Semen analysis is a central part of infertility investigation, but the clinical value in predicting male fertility is uncertain~\cite{tomlinson2016uncertainty}. 
Standard semen analysis should be performed according to the recommendations made by the \gls{who}, which includes methods of assessing semen volume, sperm concentration, total sperm count, sperm motility, sperm morphology, and sperm vitality~\cite{WHOmanual2010laboratory}. 
Sperm motility is categorized into the percentage of progressive, non-progressive, and immotile spermatozoa. Sperm morphology is classified according to the presence of head defects, neck and midpiece defects, principal piece (main part of the tail) defects, and excess residual cytoplasm in a stained preparation of cells. Figure~\ref{fig:sperm} shows an example of a frame extracted from a video of a wet human semen sample. The \gls{who} has established reference ranges for various semen parameters based on the semen quality of fertile men whose partners had a time to pregnancy up to and including $12$ months~\cite{cooper2009world}. However, these ranges can not be used to distinguish fertile from infertile men. Manual semen analysis requires trained laboratory personnel, and even when performed in agreement with the \glspl{who} guidelines, it may be prone to high intra- and inter-laboratory variability. 

Attempts to develop automatic systems for semen analysis have been carried out for several decades~\cite{mortimer2015future}. \Gls{casa} was introduced during the 1980s after the digitization of images made it possible to analyze images using a computer. 
A more rapid and objective assessment of sperm concentration and sperm motility was expected by using \gls{casa}, but it has been challenging to obtain accurate and reproducible results~\cite{mortimer2015future}. 
The results may be unreliable due to particles and other cells than spermatozoa in the sample as well as the occurrence of sperm collisions and crossing sperm trajectories. 
Better results are obtained when analyzing spermatozoa separated from seminal plasma and re-suspended in a medium. \Gls{casa} was also developed for assessment of sperm morphology and DNA fragmentation in the sperm. It is claimed that new models can also assess vitality and that some functional tests of a semen sample are possible~\cite{mortimer2015future}. However, the assessments require special staining or preparation procedures. Despite its long history as a digitized sperm analyzer, \gls{casa} is not recommended for clinical use~\cite{mortimer2015future,WHOmanual2010laboratory}. The technology, however, has been improved, and it has been suggested that using \gls{casa} for sperm counting and motility assessment can be a useful tool with less analytical variance than the manual methods~\cite{dearing2014validation, dearing2019can}.



Concerning automatic semen analysis in general, Urbano et al.~\cite{Urbano2017} present a fully automated multi-sperm tracking algorithm, which can track hundreds of individual spermatozoa simultaneously. Additionally, it is also able to measure motility parameters over time with minimal operator intervention. The method works by applying a modified version of the \gls{jpdaf} to microscopic semen recordings, allowing them to track individual spermatozoa at proximities and during head collisions (a common issue with existing \gls{casa} instruments). 
The main contribution made by Urbano et al. is the modified \gls{jpdaf} algorithm for tracking individual spermatozoa, but by only evaluating the proposed approach on two samples, the generalizability of the method to a larger population is difficult to determine. 

Dewan et al.~\cite{Dewan2018} present a similar method, tracking spermatozoa by generating trajectories of the cells across microscopic video sequences. Similar to \gls{casa}, object proposals are generated through a greyscale edge detection algorithm, which is then tracked to generate object trajectories. These trajectories are then classified into "sperm" or "non-sperm" entities using a \gls{cnn}, of which the "sperm" entities are used to estimate three quality measurements for motility (progressive, non-progressive, and immotile), and the concentration of spermatozoa per unit volume of semen. The results seem promising but since the method was evaluated on a closed dataset, it is not possible to directly compare this approach with other methods.


Although not the focus in our work, another essential attribute for semen quality is measuring the number of abnormal spermatozoa present in a semen sample. Ghasemian et al.~\cite{ghasemian2015efficient} tried to detect abnormal spermatozoa by individually classifying human spermatozoa into normal or abnormal groups. 
Shaker et al.~\cite{shaker2017dictionary} did a similar study to predict sperm heads as normal or abnormal by splitting images of sperm heads into square patches and using them as training data for a dictionary-based classifier. 
%
%
A common theme is that all automatic approaches, for both motility and morphology assessment, focus on one modality and do not incorporate other data into the analysis. Additionally, the evaluation is performed on a rather limited or closed data which hinders reproducibility and comparability of the results. In the presented work, we aim to contribute to the field of automated semen analysis in the following three ways: (i) to develop a rapid and consistent method for analyzing sperm motility automatically, (ii) to explore the potential of multimodal analysis methods combining video data with participant data to improve the results of the automatic analysis, and (iii) to compare different methods for predicting sperm motility using algorithms based on deep learning and classical machine learning.

To the best our knowledge, no study has been performed on how deep learning and multimodal data analysis may be used to directly analyze semen recordings in combination with participant/patient data for the automated prediction of motility parameters. Using data from $85$ participants and three-fold cross-validation, we observe that the initial results are promising. Thus, machine learning-based automatic analysis may become a valuable tool for the future of male infertility investigation.

\section*{Methods}\label{section:methodology}
\subsection*{Experimental Design}

Our main approach is the use of \glspl{cnn} to analyze sequences of frames from video recordings of human semen under a microscope to predict sperm motility in terms of progressive, non-progressive, and immotile spermatozoa. The video recordings are then combined with participant data to see how it may improve our methods using the multiple modalities available in our dataset. As there are no related works for which to compare directly, we first trained a series of machine learning algorithms to set a baseline for how well we can expect our deep learning-based algorithms to perform. 

The presentation of our methods is divided into three parts. Firstly, we provide a description of the dataset used for both training and evaluation of the presented methods and the statistical analysis. Secondly, we detail how we trained and evaluated the methods based on classical machine learning algorithms. Lastly, we describe our primary approach of using deep learning-based algorithms to predict sperm motility in terms of progressive, non-progressive, and immotile spermatozoa. All experiments were performed following the relevant guidelines and regulations of the Regional Committee for Medical and Health Research Ethics - South East Norway, and the \gls{gdpr}.

\begin{figure*}[t!]
\includegraphics[width=\textwidth]{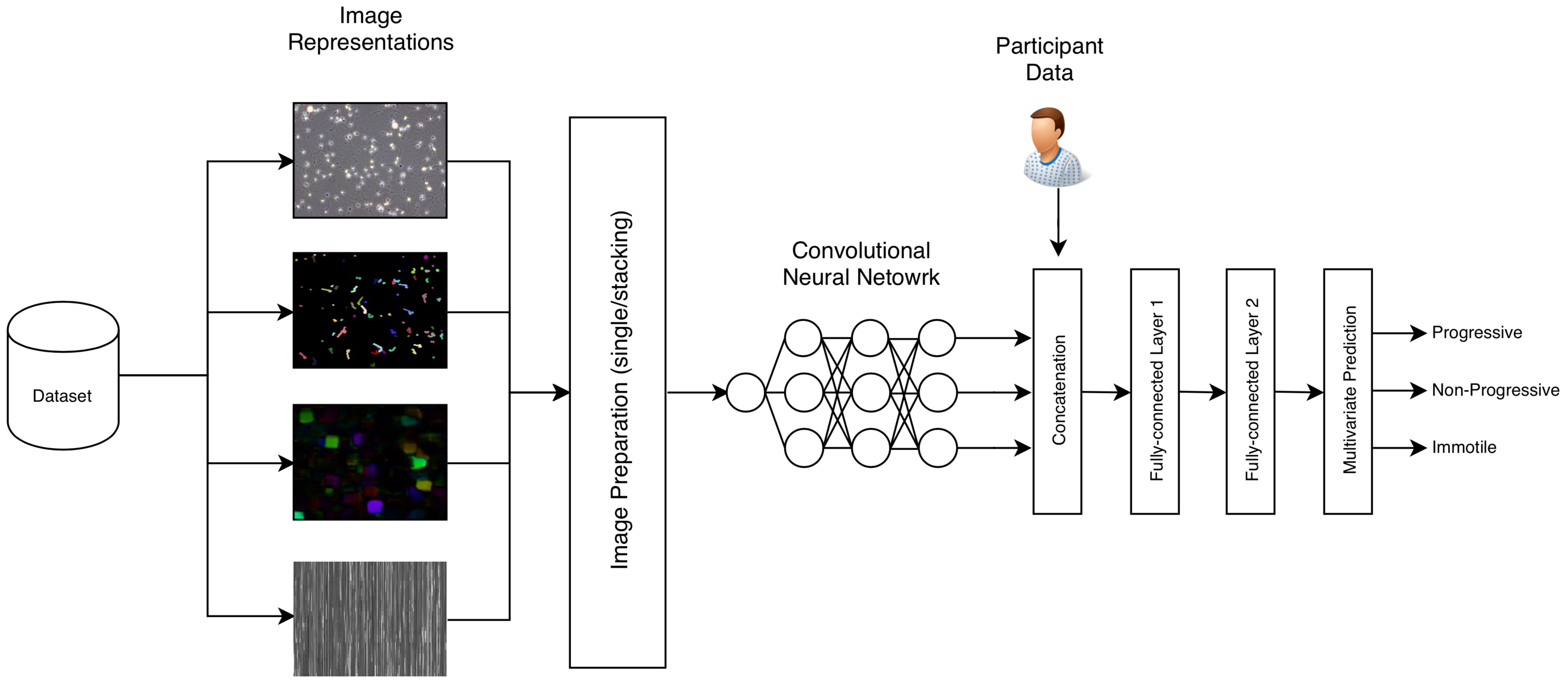}
\vspace{-2mm}

\caption{The deep learning pipeline used for all multimodal neural network-based experiments. Starting with our dataset, we extract frame data into four different representations. These four different "images" are sent to the image preparation were we either pass a single image or stacked images to a convolutional neural network (CNN). The \gls{cnn} is trained to learn a model that captures the spatial or spatial and temporal combined features of sperm motility. This is based on the image representation and preparation (stacking or single frame).
The output of the \gls{cnn} model is then combined with the participant data. This combined vector is passed through two fully-connected layers before performing multivariate prediction on the three motility variables.}
\label{figure:deep_learning_pocess}
\end{figure*}

\subsection*{Dataset}\label{subsection:dataset_details}

For all experiments, we used videos and several variables from the VISEM-dataset~\cite{visem}[https://datasets.simula.no/visem/], a fully open and multimodal dataset with anonymized data and videos of semen samples from $85$ different participants. In addition to the videos, the selected variables for the analysis included manual assessment of sperm concentration and sperm motility for each semen sample and participant data. Participant data consisted of age, \gls{bmi}, and days of sexual abstinence. 
In the experiments, the videos and participant data were used as independent variables whereas the sperm motility values (percentage of progressive, non-progressive sperm motility, and immotile spermatozoa) were used as the dependent variables. We also performed an additional experiment to test the effect of sperm concentration if added as an independent variable to the analysis. 

Details on the collection and handling of semen samples have previously been described by Andersen et al.~\cite{andersen2015body}. Briefly, the semen samples were collected at a room near the laboratory or at home and handled according to the \gls{who} guidelines~\cite{WHOmanual2010laboratory}. Samples collected at home, were transported close to the body to avoid cooling and analyzed within two hours. Assessment of sperm concentration and sperm motility was performed as described in the \gls{who} 2010 manual \cite{WHOmanual2010laboratory}. Sperm motility was evaluated using videos of the semen sample, and all samples were assessed by one experienced laboratory technician. $10$~$\mu l$ of semen were placed on a glass slide, covered with a $22\times22$ mm cover slip and placed under the microscope. Videos were recorded using an Olympus CX31 microscope with phase contrast optics, heated stage (37~\degree C), and a microscope mounted camera (UEye UI-2210C, IDS Imaging Development Systems, Germany). Videos for sperm motility assessment were captured using $400\times$~magnification and stored as AVI files. The recordings vary in length between two to seven minutes with a frame rate of $50$ frames-per-second.


\subsection*{Statistical Analysis}
For all experiments, we report the \gls{mae} calculated over three-fold cross-validation to get a more robust and generalizable evaluation. Furthermore, statistical significance was tested by a corrected paired t-test, where a p-value below or equal to $0.05$ was considered significant. Usually, t-test is based on the assumption that samples are independent. However, samples in the folds of cross-validation are not independent. Therefore, a fudge factor is needed to compensate for the not independent samples~\cite{nadeau2000inference}. The significance test showed that all results with an average \gls{mae} below $11$ are significant improvements compared to the \gls{zeror} baseline. For \gls{zeror}, which is also commonly known as the null model, the cross-validation coefficient is defined with a $Q2$ value of $0$. This means that the \gls{zeror} predictions are equal to the average calculated over the entire training dataset.

\subsection*{Baseline Machine Learning Approach}\label{section:machine_learning}
For the machine learning baseline, we relied on a combination of well-known algorithms and handcrafted features. To extract features from the video frames, we used the open-source library Lucene Image Retrieval (LIRE)~\cite{lux2016lire}. LIRE is a Java library that offers a simple way to retrieve images and photos based on color and texture characteristics. We tested all available features (more than $30$ different ones) with all machine learning algorithms (more than $40$ different ones), but in this work, we only report the features that worked best with our machine learning algorithms, which were the Tamura features. Tamura features (coarseness, contrast, directionality, line-likeness, regularity, and roughness) are based on human visual perception, which makes them very important in image representation. Using the Tamura image features, participant data and a combination of both, we trained different algorithms to perform prediction on the motility variables. We performed a total of three experiments per tested algorithm; one using only Tamura features, one using only participant data, and one combining the Tamura features with the participant data through early fusion. 

Since the Tamura features are sparse compared to deep features, we used a slightly different approach for selecting frames from the videos. Each video was represented by a feature vector containing the Tamura features of two frames per second (the first and the middle frame) for the first $60$ seconds. In total, we had $120$ frames per video and a visual feature space consisting of $2160$ feature points. These features were then used to train multiple machine learning algorithms using the WEKA machine learning library~\cite{Weka2009}. We conducted experiments with all available algorithms, but report only the six best performing ones. The reported algorithms are Simple Linear Regression, Random Forests, Gaussian Process, Sequential Minimal Optimization Regression (SMOreg), Elastic Net, and Random Trees. One limitation of these algorithms is that they are only able to predict one value at a time, meaning we had to run them once for each of the three sperm motility variables. 


\subsection*{Deep Learning Approach}
For our primary approach, we use methods based on \glspl{cnn} to perform regression on the three motility variables. For each deep learning-based experiment, we extracted $250$ frame samples (single frames or frame sequences) from each of the $85$ videos of our dataset. The reason for only extracting $250$ frames per video was due to some videos being too short for collecting more than $250$ sequences of $30$ frames, which is about $7,500$ frames equalling about $2$ minutes of video at $50$ frames-per-second. 
This results in a total of $21,250$ frames used for training and validation. As we are evaluating each method using three-fold cross-validation, the split between the training and validation datasets is $14,166$ and $7,083$ frame samples, respectively.

Our deep learning approaches can be split into three groups. Firstly, we analyze raw frames as they are extracted from the videos. The analysis is done by looking at the raw pixel values from a single or a sequence of frames and using these to make a prediction. 
Secondly, we use optical flow to generate temporal representations of frame sequences to condense the information of the temporal dimension into a single image. The advantages of this representation is that it can model the temporal dependencies in the videos, and it is able to alleviate the hardware costs of analyzing raw frame sequences using \glspl{cnn}. 
Lastly, we combine the two previous methods to exploit the advantages of both, by using the visual features of raw video frames together with the temporal information of the optical flow representations. 

The baseline for the deep learning approaches are the machine learning algorithms as described above and \gls{zeror}. For each experiment, we predict the percentage of progressive spermatozoa, non-progressive spermatozoa, and immotile spermatozoa for a single semen sample. In contrast to the classical machine learning algorithms, neural networks can predict all three values at once. Figure~\ref{figure:deep_learning_pocess} illustrates a high level overview of the complete deep learning analysis pipeline.

All deep learning-based models were trained using \gls{mse} to calculate loss and Nadam~\cite{Dozat2015IncorporatingNM} to optimize the weights. The Nadam optimizer had a learning rate of $0.002$, $\beta_1$ value of $0.900$, and $\beta_2$ value of $0.999$. We trained each model for as long as it improved with a patience value of $20$ epochs, meaning if the \gls{mse} did not improve on the validation set for $20$ epochs, we stopped the training to avoid overfitting. The model used for evaluation was the one which performed best on the validation set, not the model from the last epoch. Furthermore, for each method we trained two models. One model uses only frame data, and the other uses a combination of the frame data and the related participant data (\gls{bmi}, age, and days of sexual abstinence). To include the participant data in the analysis, we first pass a frame sample through the \gls{cnn}. Then, we take the output of the last convolutional layer and globally average pool it to produce a one-dimensional feature vector which is concatenated with the participant data. This combined vector is then passed through two fully-connected layers consisting of $2,048$ neurons each before being making the final prediction (shown in Figure~\ref{figure:deep_learning_pocess}). In the following few sections, we will describe six different methods used to predict sperm motility; a method using single frames for prediction, a method which stacks frames channel-wise, a method using vertical frame matrices, a method based on sparse optical flow, a method based on dense optical flow, and a method based on two-stream networks.


\subsubsection*{Single Frame Prediction}\label{subsubsection:single}
For the single frame-based method, we extracted $250$ single frames from each video and used this to train various \glspl{cnn} models based on popular neural network architectures (such as DenseNet~\cite{huang2017densely}, ResNet~\cite{He2016DeepRL}, and Inception~\cite{szegedy2015rethinking}). We experimented using transfer learning from the ImageNet~\cite{imagenetcvpr09} weights included with the Keras~\cite{chollet2015keras} implementations of the different \gls{cnn} architectures and found that, in general, using these weights as a base for further training worked better than training from scratch. Note that we did not fine-tune the models, meaning we did not freeze any layers during training. We only report the model which performed best, which in our case was a ResNet-50 model implemented in Keras with a TensorFlow~\cite{Tensorflow2015} back-end. The frames were resized to $224\times224$ before being passed through the model, which is the recommended size for the ResNet-based architectures~\cite{He2016DeepRL}.

The single frame-based approach is simple and comes with some obvious limitations. Most notably, we lose the temporal information present within the video. Losing the temporal information may be acceptable when measuring attributes that rely on visual clues, such as morphology, but for motility the change over time is an important feature.

\begin{figure}[t!]
\includegraphics[width=\textwidth]{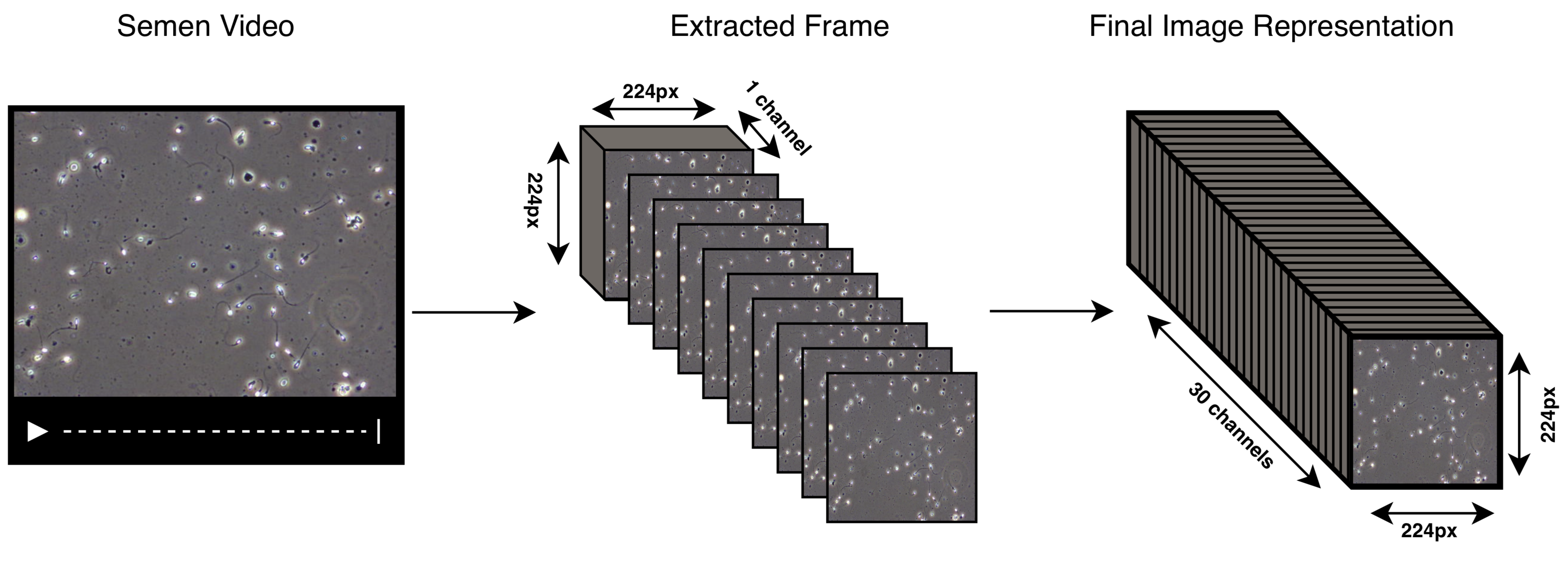}
\caption{An illustration of how frames are stacked channel-wise after being greyscaled. From a video, a sequence of \textit{n} frames are extracted and greyscaled. These frames are then stacked channel-wise, meaning each frame occupied one channel-dimension of the final image. The final stacked "image" is then of shape $224\times224\times30$.}
\label{figure:frame_stacking_example}
\end{figure}

\subsubsection*{Greyscale Frame Stacking}\label{subsubsection:greyscale_stacking}
The Grayscale Frame Stacking method is an extension of the single-frame prediction approach. Here, we extract $250$ batches of $30$ frames and greyscaled them before stacking them channel-wise (shown in Figure~\ref{figure:frame_stacking_example}). This results in $21,250$ frame samples with a shape of $224\times224\times30$, which contains the information of $30$ consecutive frames. The reasons for greyscaling the frames before stacking them is two-fold. Firstly, seeing as the color of the videos are a feature of the microscope and lab preparation, and not the spermatozoon itself, we assume that this feature may confuse the model in unintended ways. Secondly, greyscaling the frames reduces the size of each frame by three, making stacking $30$ frames feasible on less powerful hardware. The motivation behind this approach was to keep the temporal information present in a given frame sequence, yet still, keep the size of the input relatively small.

These extracted frame sequences were used to train a ResNet-50 model implemented in Keras~\cite{chollet2015keras}. Note that because we changed the size of the channel dimension, we could not perform transfer learning as we did in the previous method. Apart from this, the model was trained in the same manner as described in the beginning of the Deep Learning Approach section.

\begin{figure*}[!t]
    \newcommand{\figsize}{3.3cm}
    \centering
    \setlength\tabcolsep{1.5pt}
    \renewcommand{\arraystretch}{1}
    \begin{tabular}{m{10pt}m{\figsize}m{\figsize}m{\figsize}m{\figsize}m{\figsize}m{1pt}}
    
    \begin{tabular}{c} \vspace{-26px}\textbf{1} \end{tabular} &
    \subfloat[ ]{%
        \centering
        \includegraphics[width=\figsize, height=\figsize]{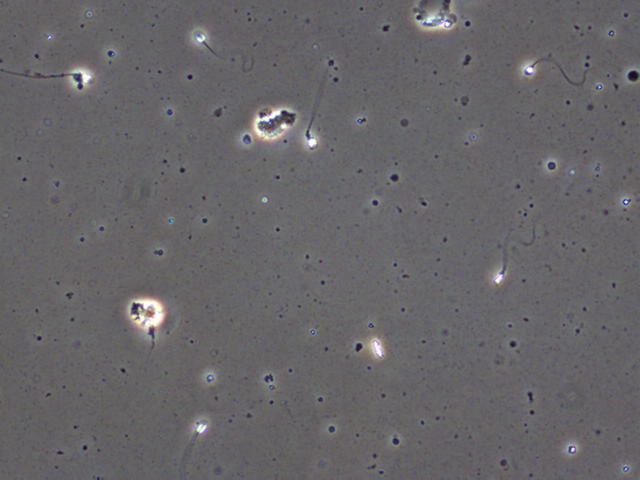}
    } &
    \subfloat[ ]{%
        \centering
        \includegraphics[width=\figsize, height=\figsize]{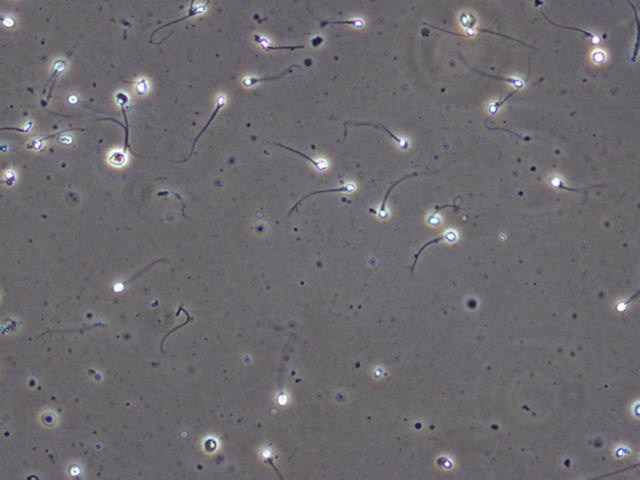}
    } &
    \subfloat[ ]{%
        \centering
        \includegraphics[width=\figsize, height=\figsize]{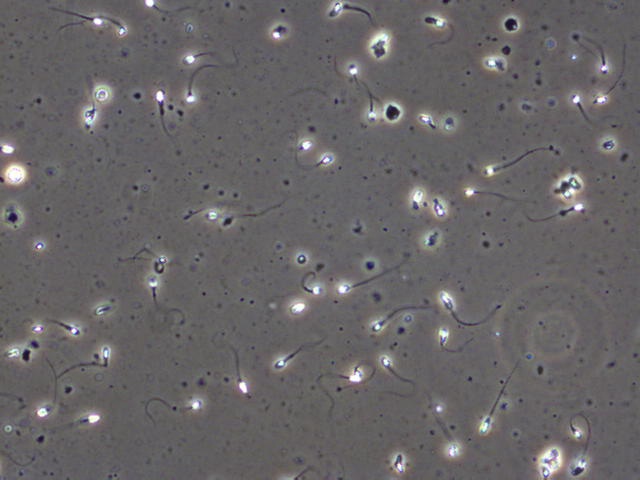}
    } &
    \subfloat[ ]{%
        \centering
        \includegraphics[width=\figsize, height=\figsize]{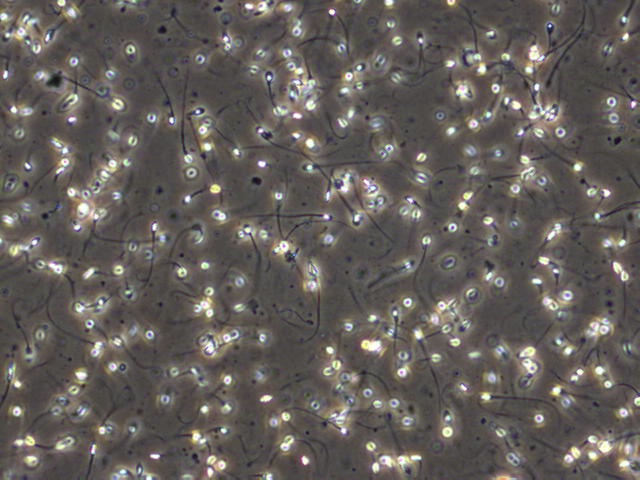}
    } &
    \subfloat[ ]{%
        \centering
        \includegraphics[width=\figsize, height=\figsize]{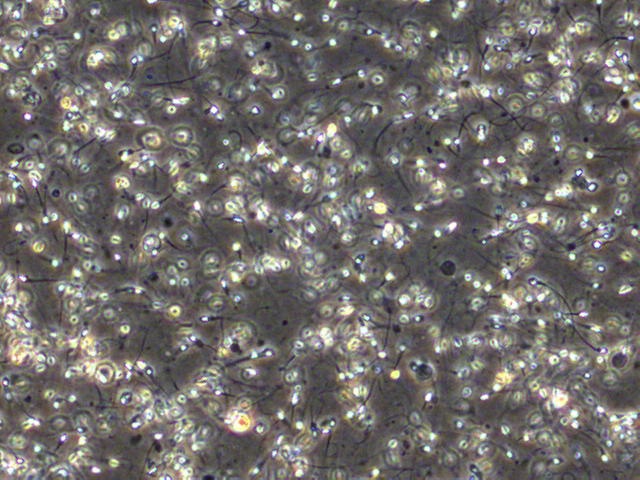}
    } & \\
    \begin{tabular}{c} \vspace{-13px}\textbf{2} \end{tabular} &
    \subfloat{%
        \centering
        \includegraphics[width=\figsize, height=\figsize]{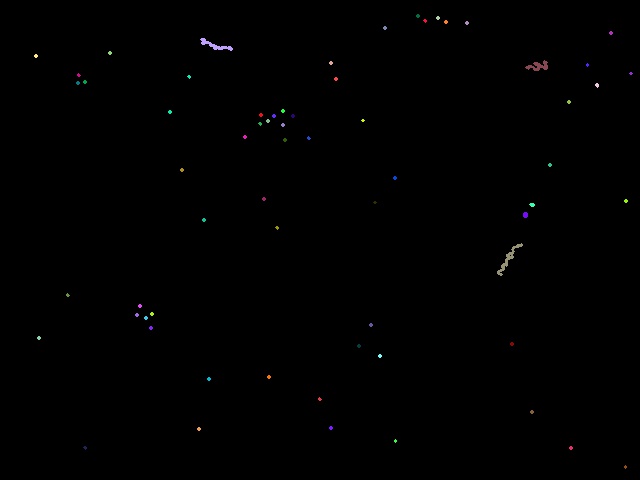}
    } &
    \subfloat{%
        \centering
        \includegraphics[width=\figsize, height=\figsize]{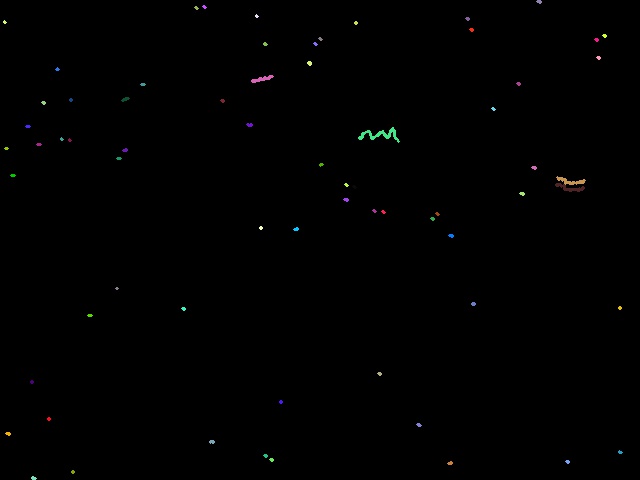}
    } &
    \subfloat{%
        \centering
        \includegraphics[width=\figsize, height=\figsize]{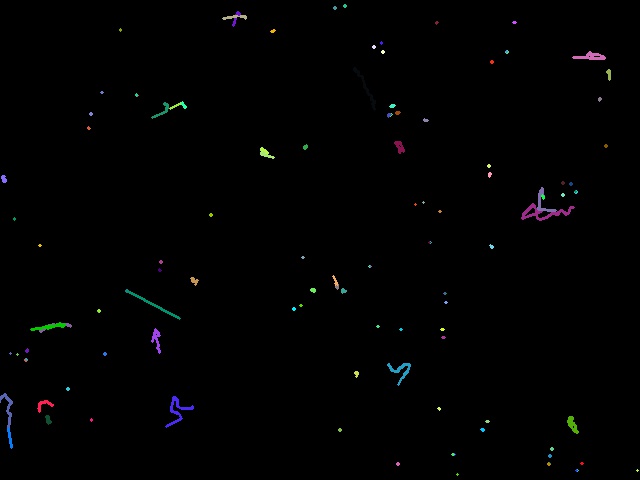}
    } &
    \subfloat{%
        \centering
        \includegraphics[width=\figsize, height=\figsize]{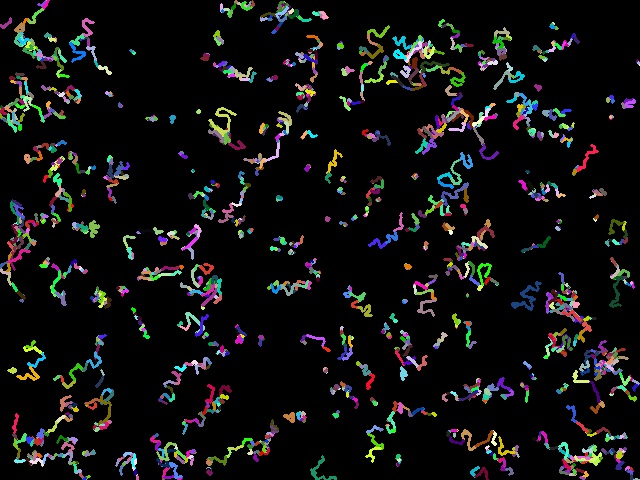}
    } &
    \subfloat{%
        \centering
        \includegraphics[width=\figsize, height=\figsize]{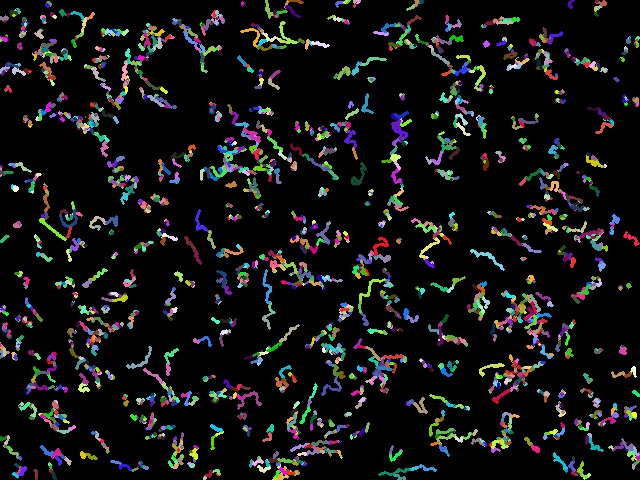}
    } & \\
    \begin{tabular}{c} \vspace{-13px}\textbf{3} \end{tabular} &
    \subfloat{%
        \centering
        \includegraphics[width=\figsize, height=\figsize]{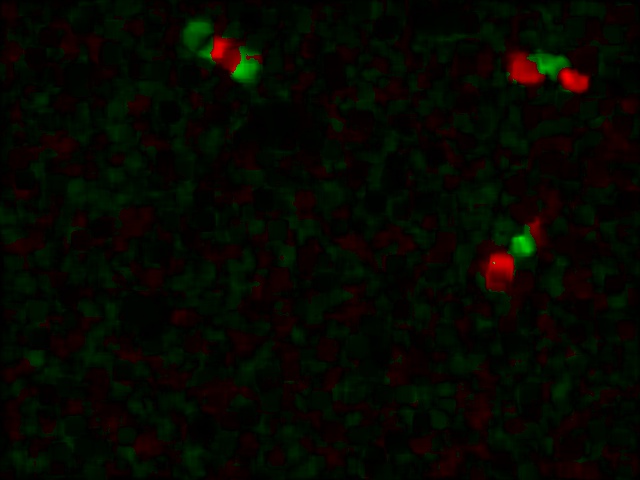}
    } &
    \subfloat{%
        \centering
        \includegraphics[width=\figsize, height=\figsize]{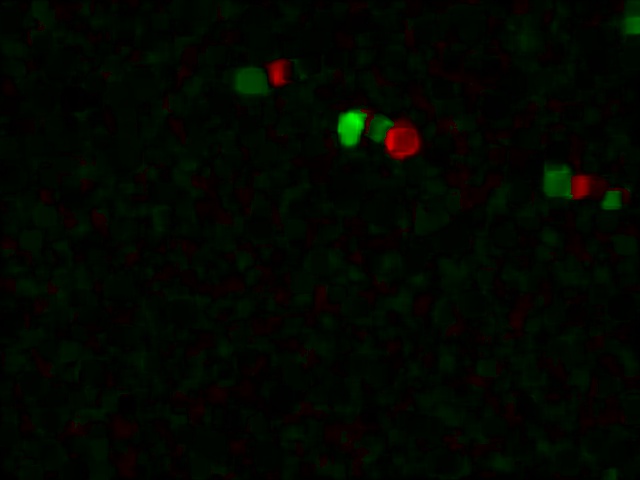}
    } &
    \subfloat{%
        \centering
        \includegraphics[width=\figsize, height=\figsize]{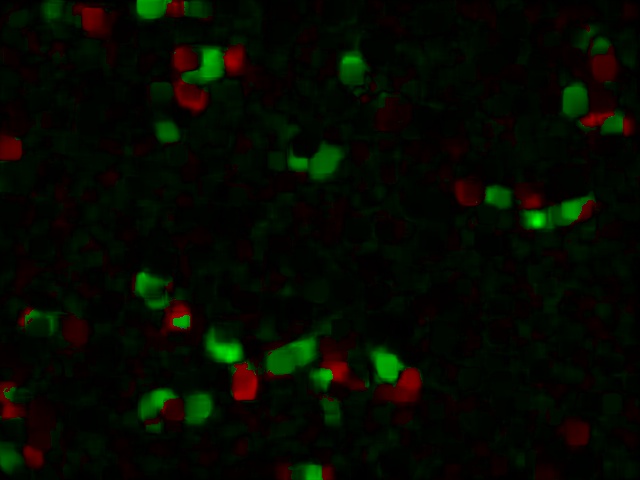}
    } &
    \subfloat{%
        \centering
        \includegraphics[width=\figsize, height=\figsize]{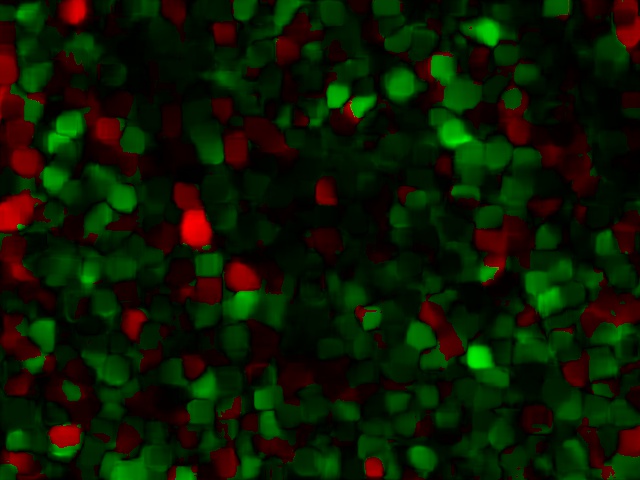}
    } &
    \subfloat{%
        \centering
        \includegraphics[width=\figsize, height=\figsize]{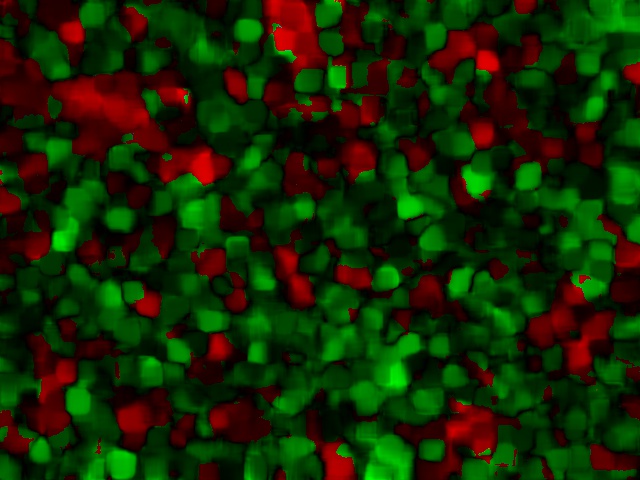}
    } & \\
    \begin{tabular}{c} \vspace{-13px}\textbf{4} \end{tabular} &
    \subfloat{%
        \centering
        \includegraphics[width=\figsize, height=\figsize]{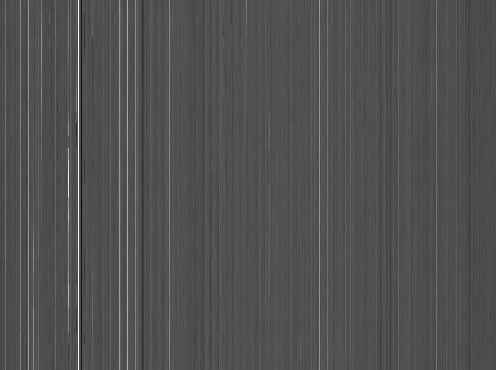}
        \label{subfig:LK_low_example}
    } &
    \subfloat{%
        \centering
        \includegraphics[width=\figsize, height=\figsize]{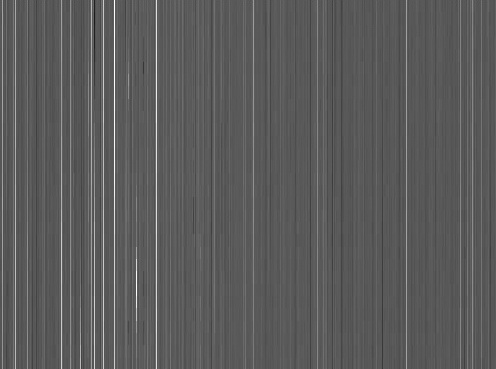}
        \label{subfig:LK_low_mid_example}
    } &
    \subfloat{%
        \centering
        \includegraphics[width=\figsize, height=\figsize]{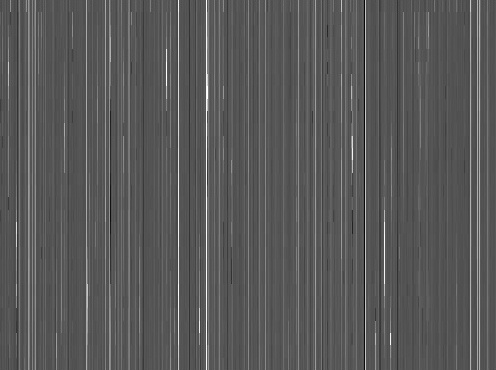}
        \label{subfig:LK_mid_example}
    } &
    \subfloat{%
        \centering
        \includegraphics[width=\figsize, height=\figsize]{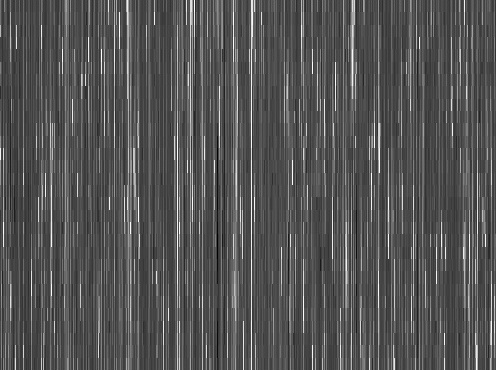}
        \label{subfig:LK_mid_high_example}
    } &
    \subfloat{%
        \centering
        \includegraphics[width=\figsize, height=\figsize]{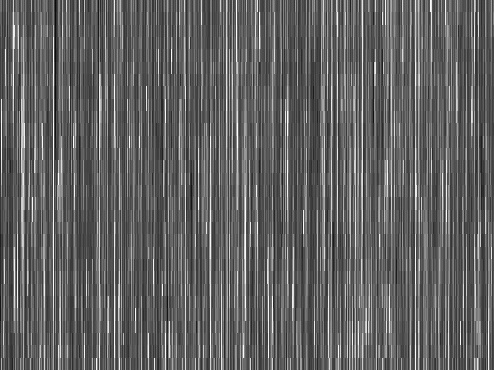}
        \label{subfig:LK_high_example}
    } &
    
    \end{tabular}

    \vspace{-2mm}
    \caption{Examples of images from videos of semen samples with different concentrations (columns) and the four image representations used to train the neural network-based algorithms (rows). Sperm concentration; A) $4$ per $x10^6/mL$, B) $33$ per $x10^6/mL$, C) $105$ per $x10^6/mL$, D) $192$ per $x10^6/mL$, and E) $350$ per $x10^6/mL$. Image representation; 1) original video, 2) sparse optical flow, 3) dense optical flow, and 4) vertical frame matrix.}\label{fig:layer_representations}
\end{figure*}

\subsubsection*{Vertical Frame Matrix}
To create the vertical frame matrix, $250$ batches of $30$ frames were extracted and greyscaled. Each frame was resized to $64\times64$ before being flattened into a one-dimensional vector. The reason for resizing each frame was to keep the length of the flattened images relatively short. With a size of $64\times64$, the final vector had a length of $4096$. Each vector was then stacked on top of each other which resulted in a matrix with a shape of $30\times4096\times1$. Examples images using this transformation can be seen in row four of Figure~\ref{fig:layer_representations}. Similar to the Greyscale Frame Stacking approach described in the previous section, we condense the information of multiple frames into a single image, which we can then pass through a standard two-dimensional \gls{cnn}. Due to size constraints, the model used for this method was ResNet-18. Otherwise, it was trained in the same way as the previous two methods. 


\subsubsection*{Sparse Optical Flow}
For the Sparse Optical Flow approach, we use Lucas-Kanade's~\cite{Lucas1981} algorithm of estimating optical flow. What makes sparse optical flow "sparse," is that we only measure the difference between a few tracked features from one frame to another. In our case, we use Harris and Stephens corner detection algorithm~\cite{Harris88acombined} to detect individual sperm heads (implemented in OpenCV~\cite{opencv_library} as "goodFeaturesToTrack"). Then, we track the progression of each spermatozoon using Lucas-Kanade's algorithm over a sequence of $30$ frames. Similar to the previous methods, sequences were sampled at evenly spaced intervals to maximize differences between optical flow representations. We used a \gls{cnn} model based on the ResNet-50 architecture implemented in Keras and trained using the same configuration described previously.
Examples for the sparse optical flow image representation can be seen in row two of Figure~\ref{fig:layer_representations}.

\begin{table}[t]
    \def\arraystretch{1.1}
    \setlength\tabcolsep{3pt}
    
    \centering
    Classical Machine Learning Results
    
    \par\bigskip
    \centering
        \resizebox{.6\textwidth}{!}{%
            \begin{tabular}{ l | c c c | c } 
                \bottomrule
                Method & Progressive & Non-progressive & Immotile & Average Mean Absolute Error   \\ \toprule
                \multicolumn{5}{c}{Baseline} \\
                \bottomrule
                ZeroR & 17.260 & 7.860 & 13.660 & 12.927 \\ \hline
                \toprule
                \multicolumn{5}{c}{Participant Data Only} \\
                \bottomrule
                Elastic Net & 15.198 & 9.525 & 13.441 & 12.721 \\ \hline
                Gaussian Process & 15.556 & 9.762 & 13.474 & 12.931 \\ \hline
                Simple Linear Regression & 15.416 & 9.281 & 13.601 & 12.766 \\ \hline
                SMOreg & 15.355 & 9.441 & 12.959 & 12.585 \\ \hline
                \textbf{Random Forests} & \textbf{13.312} & \textbf{8.886} & \textbf{11.905} & \textbf{11.368} \\ \hline
                Random Tree & 17.801 & 10.952 & 14.984 & 14.579 \\ \hline
                \toprule
                \multicolumn{5}{c}{Tamura Image Features Only} \\
                \bottomrule
                Elastic Net & 14.400 & 7.750 & 12.190 & 11.447 \\ \hline
                Gaussian Process & 13.230 & 7.260 & 11.920 & 10.803 \\ \hline
                Simple Linear Regression & 13.520 & 8.170 & 12.690 & 11.460 \\ \hline
                \textbf{SMOreg} & \textbf{13.220} & \textbf{7.260} & \textbf{11.920} & \textbf{10.800} \\ \hline
                Random Forests & 13.530 & 7.400 & 12.060 & 10.997 \\ \hline
                Random Tree & 18.700 & 9.960 & 16.520 & 15.060 \\ \hline
                \toprule
                \multicolumn{5}{c}{Tamura Image Features and Participant Data} \\
                \bottomrule
                Elastic Net & 14.130 & 9.890 & 11.750 & 11.923 \\ \hline
                Gaussian Process & 13.700 & 10.120 & 11.460 & 11.760 \\ \hline
                Simple Linear Regression & 13.940 & 10.240 & 11.410 & 11.863 \\ \hline
                SMOreg & 13.710 & 10.140 & 11.460 & 11.770 \\ \hline
                \textbf{Random Forests} & \textbf{13.510} & \textbf{10.000} & \textbf{11.340} & \textbf{11.617} \\ \hline
                Random Tree & 18.660 & 13.270 & 16.960 & 16.297 \\
                
                \toprule
            \end{tabular}%
        }
    \caption{Prediction performance of the machine learning-based methods in terms of mean absolute error for each of the motility values and the overall average. The best performing algorithm in each category is in bold.}
    \label{table:machine_learning_results}
\end{table}									

\subsubsection*{Dense Optical Flow}
The Dense Optical Flow approach generates optical flow representations using Gunner Farneback's algorithm~\cite{Gunnar2003, simonyan2014two} for two-frame motion estimation. Dense optical flow, in contrast to sparse optical flow, processes all pixels of a given image instead of a few tracked features. For this method, we tried two configurations. The first configuration measures the difference between two consecutive frames. The second configuration adds a stride of $10$ frames between selected frame samples. This is done to increase the measured difference between frame comparisons. We collected $250$ dense optical flow images and trained one model for each of the two configurations to evaluate the result of this method. For both stride configurations, we train each model using the same architecture (ResNet-50) and training configuration as for the other deep learning methods.
Examples for the created image representations using the dense optical flow can be seen in row three of Figure~\ref{fig:layer_representations}.

\begin{table}[t]
    \def\arraystretch{1.1}
    \setlength\tabcolsep{3pt}
    \centering
    Deep Learning Results
    \par\bigskip
    \centering
        \resizebox{.6\textwidth}{!}{%
            \begin{tabular}{ l | c c c | c } 
                \bottomrule
                Method & Progressive & Non-progressive & Immotile & Average Mean Absolute Error   \\
                \toprule
                \multicolumn{5}{c}{Raw Frame Data Approach} \\
                \bottomrule
                Single Frames (ResNet50) & 13.162 & 8.024 & 10.967 & 10.718 \\ \hline
                Single Frames (ResNet50) + PD & 13.659 & 8.196 & 12.293 & 11.383 \\\hline
                \textbf{Channel-wise Greyscale} & \textbf{10.498} & \textbf{7.037} & \textbf{8.822} & \textbf{8.786} \\ \hline
                Channel-wise Greyscale + PD & 11.599 & 7.849 & 10.132 & 9.860 \\ \hline
                Vertical Frame Matrix & 11.149 & 8.218 & 9.418 & 9.595 \\ \hline
                Vertical Frame Matrix + PD & 11.182 & 8.199 & 9.274 & 9.552 \\ 
                \toprule
                \multicolumn{5}{c}{Optical Flow Approach} \\
                \bottomrule
                Sparse Optical Flow & 11.573 & 7.263 & 10.155 & 9.664 \\ \hline
                Sparse Optical Flow + PD & 12.214 & 7.760 & 10.802 & 10.259 \\ \hline
                \textbf{Dense Optical Flow (stride=1)} & \textbf{10.191} & \textbf{7.114} & \textbf{8.914} & \textbf{8.740} \\ \hline
                Dense Optical Flow (stride=1) + PD & 10.795 & 7.856 & 8.745 & 9.132 \\ \hline
                Dense Optical Flow (stride=10) & 10.319 & 7.546 & 8.782 & 8.882 \\ \hline
                Dense Optical Flow (stride=10) + PD & 11.386 & 7.825 & 9.734 & 9.648 \\
                \toprule
                \multicolumn{5}{c}{Two Stream Network Approach} \\
                \bottomrule
                Two Stream Sparse & 15.888 & 8.187 & 13.326 & 12.467 \\ \hline
                Two Stream Sparse + PD & 16.435 & 8.197 & 13.172 & 12.601 \\ \hline
                Two Stream Dense (stride=1) & 14.583 & 7.393 & 11.996 & 11.324 \\ \hline
                Two Stream Dense (stride=1)+ PD & 18.166 & 8.570 & 15.983 & 13.940 \\ \hline
                \textbf{Two Stream SP + DE (stride=1)} & \textbf{11.848} & \textbf{7.070} & \textbf{10.823} & \textbf{9.917} \\ \hline
                Two Stream SP + DE (stride=1) + PD & 17.304 & 8.066 & 13.783 & 13.051 \\ \hline
                \toprule
            \end{tabular}%
        }
    \caption{Prediction performance of the deep learning-based methods in terms of mean absolute error for each of the motility values and overall mean. Note that for each method, we trained two models, one with participant data and one without. Methods which used participant data under training are marked with (+ PD). For the methods which use dense optical flow, stride represents the number of frames skipped when comparing the difference of two frames.}
    \label{table:deep_learning_results}
\end{table}

\begin{figure*}[t!]
\begin{tikzpicture}
\begin{axis}[
    enlarge x limits=0.03,
    every axis plot post/.style={/pgf/number format/fixed},
    ybar=1pt, 
    bar width=10pt,
    legend style={at={(0.03,.8)},anchor=west},
    x=1cm,
    ymin=8,
    ymax=18,
    axis on top,
    xtick=data,
    xticklabel style={, text width=.8cm, font=\scriptsize, align=center},
    visualization depends on=rawy\as\rawy,
    after end axis/.code={
        \draw [ultra thick, white, decoration={snake, amplitude=2pt}, decorate] (rel axis cs:0,1.05) -- (rel axis cs:1,1.05);
    },
    every node near coord/.append style={
        font=\tiny,
        inner sep=1pt,rotate=90,xshift=0.3cm
    },
    axis lines*=left,
    clip=false,
    symbolic x coords={
        Single Frame,
        Channel-wise Greyscale,
        Vertical Frame Matrix, 
        Sparse Optical Flow,
        Dense Optical Flow~(1),
        Dense Optical Flow~(10),
        Two Stream Sparse,
        Two Stream Dense,                
        Two Stream Dense + Sparse,
        ZeroR,
        Elastic Net,
        Gaussian Process,
        Simple Linear Regression,
        SMOreg,
        Random Forests,
        Random Tree
    }
]

\addplot coordinates {
    (Single Frame,10.7176) (Channel-wise Greyscale,8.7856) (Vertical Frame Matrix,9.5950) (Sparse Optical Flow,9.6635) (Dense Optical Flow~(1),8.7398)
    (Dense Optical Flow~(10),8.8824) (Two Stream Sparse,12.4669) (Two Stream Dense,11.3238)
    (Two Stream Dense + Sparse,9.9168) (ZeroR,12.9267) (Elastic Net,11.4467) (Gaussian Process,10.8033)
    (Simple Linear Regression,11.4600) (SMOreg,10.8000) (Random Forests,10.9967) (Random Tree,15.0600)
}; \addlegendentry{Without participant data}

\addplot coordinates {
    (Single Frame,11.3825) (Channel-wise Greyscale,9.8602) (Vertical Frame Matrix,9.5516)
    (Sparse Optical Flow,10.2587) (Dense Optical Flow~(1),9.1317) (Dense Optical Flow~(10),9.6481) (Two Stream Sparse,12.6012) (Two Stream Dense,13.9398) (Two Stream Dense + Sparse,13.0510)
    (ZeroR,12.9267) (Elastic Net,11.9233) (Gaussian Process,11.7600) (Simple Linear Regression,11.8633)
    (SMOreg,11.7700) (Random Forests,11.6167) (Random Tree,16.2967)
}; \addlegendentry{With participant data}

\addlegendimage{line legend}
\addlegendentry{Significance threshold compared to ZeroR baseline}



\coordinate (B) at (axis cs:Single Frame,11.0000);
\coordinate (O1) at (rel axis cs:0,0);
\coordinate (O2) at (rel axis cs:1,0);



\draw [black,sharp plot, dashed] (B -| O1) -- (B -| O2);

\end{axis}
\end{tikzpicture}
\vspace{-3mm}
\caption{The different machine learning-based algorithms (classical and deep learning) used to predict semen quality in terms of progressive, non-progressive, and immotile spermatozoon. The stippled line represents the threshold for the results to be considered significant compared to the ZeroR baseline. The y-axis does not start at 0 to better highlight the differences. For the methods which used dense optical flow, stride values, how many frames are skipped when comparing two frames, are presented with a \textit{1} or \textit{10} indicating the number of skipped frames. Dense Optical Flow (1) and Channel-wise Greyscale are the best-performing ones but, several of our proposed methods are below the significance threshold.}

\label{figure:deep_learning_results_plot}
\end{figure*}
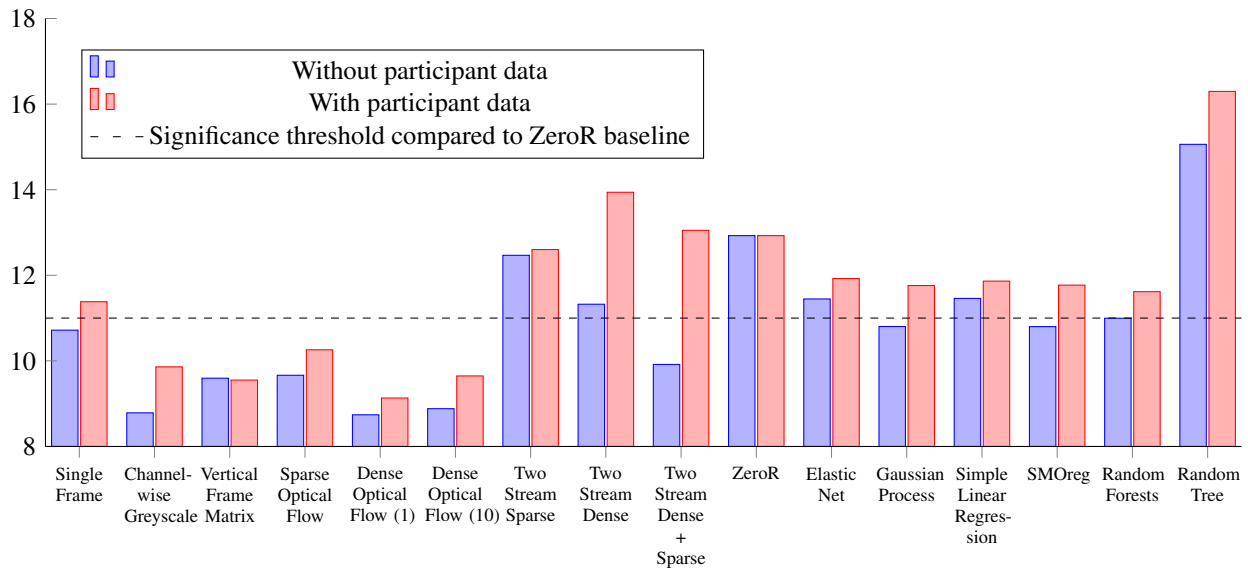

\subsection*{Two-stream Network}
For the last approach, we combine the two previous methods (visual features of raw frames and the temporal information of optical flow), which is inspired by the work done by Simonyan and Zisserman's~\cite{simonyan2014two}, where they used a dual-network to perform human action recognition and classification. The model architecture follows a similar structure as described in their article, with the difference being how we input the optical flow representations into the model (we do not stack multiple optical flow representations for different sequences).

Based on this modification, we propose three different methods. 
Firstly, we use the dual network to analyze one raw video frame in parallel with a Lukas-Kanade sparse optical flow representation of the previous $30$ frames. 
Secondly, we process one raw frame together with a Farneback's dense optical representation. 
Lastly, we again use one raw frame, but now we combine both the Lukas-Kanade and Farneback's optical flow method by stacking them channel-wise and pass these together through the network. 
Frames were extracted in the same way as performed for the Single Frame Prediction approach, and the optical flow representations were reused from the Optical Flow-based experiments. 

\section*{Results and Discussion}\label{section:results}
A complete overview of the results for each method can be seen in Table~\ref{table:machine_learning_results} and Table~\ref{table:deep_learning_results}. A chart comparing the results is presented in Figure~\ref{figure:deep_learning_results_plot}. 
Table~\ref{table:machine_learning_results} presents the results for the classical machine learning algorithms trained on participant data, Tamura image features, and a combination of the two. For these results, the Gaussian Process, SMOreg, and Random Forests have a \gls{mae} below $11$, which according to the paired t-test analysis is significant. One interesting finding is that for all cases where participant data is added, the algorithm performs worse. Although a preliminary result, for BMI this is not in line with the finding in our previous work Andersen et al.~\cite{andersen2015body}, where \gls{bmi} was found to be negatively correlated with sperm motility using multiple linear regression. However, the methods are very different and therefore not directly comparable. As future work, we plan to perform an extensive analysis of all methodologies on a new dataset. Another interesting insight gained from this experiment is that the Tamura features seem to be well suited for sperm analysis, which will be interesting to investigate more closely.  

Since sperm concentration is an important confounding variable when assessing sperm motility by \gls{casa}, we performed additional experiments using the two best-performing algorithms to investigate whether or not it had any influence. For the Random Forest, we achieved a \gls{mae} of $11.091$ when including sperm concentration, compared to $10.996$ when we did not. For SMOReg, the \gls{mae} was $10.902$ with and $10.800$ without. This minor difference in error indicates that our method is not gaining or losing any predictive power when including sperm concentration in the analysis, which can be seen as an advantage compared with CASA systems.

To assess the performance of the deep learning-based methods, we used the best performing classical machine learning approach (SMOreg with a \gls{mae} of $10.800$) and \gls{zeror} as a baseline.
%
In Table~\ref{table:deep_learning_results}, the results for single and multimodal deep learning approaches are shown. For most of the experiments, the deep learning models outperform the best machine learning algorithm (SMOreg) by a margin of one or two points. The two methods which are not significant better than \gls{zeror} are the two-stream neural networks, which combined the two optical flow representations in a custom network.

We hypothesize that this is related to the fact that these networks are not able to learn the association between the temporal information of the optical flow and the visual data of the raw frame. Similar to the machine learning algorithms, all methods which combined the participant data with the videos performed worse than those without, leading to the same conclusion as previously discussed. Thus, in our study, adding patient data does not improve the results compared to using only video data, regardless of the algorithms used. If these findings also apply to other patient data needs to be further investigated.

The best performing approaches were a near tie between the method Channel-wise Greyscale and Dense Optical Flow using a stride of 1 or 10 (see Figure~\ref{figure:deep_learning_results_plot}). The Channel-wise Greyscale approach achieved a \gls{mae} of $8.786$, which is two points lower than that of the best performing classical machine learning algorithm (see Table~\ref{table:deep_learning_results}). The two Dense Optical Flow methods have the same performance as the Channel-wise Greyscale approach but using one-tenth of the image size, which makes them faster and less computational resource demanding.

It is important to point out that the 250 frames used in the analysis were extracted evenly distributed across the entire video length. This means that if there were a noticeable reduction in sperm motility after a certain amount of time, it would be taken into account by the algorithm. The results also support this assumption as the deep learning methods outperformed all classical machine learning methods. This is one of the advantages of the deep learning-based methods presented here. 

In terms of time needed for the analysis, all presented methods perform the prediction within five minutes, including data preparation which takes most of the time. This is considerably faster than manual sperm motility assessment would be. The classical machine learning methods are faster to train, but in terms of application of the model, the speed is comparable with the deep learning methods. 

%

\section*{Conclusion and Future Work}\label{section:conclusion}
Overall, our results indicate that deep learning algorithms have the potential to predict sperm motility consistently and time efficiently. Multimodal analysis methods combining video data with participant data did not improve the prediction of sperm motility compared to using only the video data. However, it is possible that multimodal analysis using other participant data could improve the prediction. Our results indicate that the deep learning models can incorporate time into their analysis, and therefore are able to predict motility values better than the classical machine learning algorithms. In the future, deep learning-based methods could be used as an efficient support tool for human semen analysis. The presented methods can easily be applied to other relevant assessments such as automatic evaluation of sperm morphology. 

Efficient analysis of long videos is a challenge, and future work should focus on how to combine the different modalities of time, imaging, and patient data. The dataset used in this study is also shared openly to ensure comparability and reproducibility of the results. Furthermore, we hope that the methods described in this work will inspire to further development of automatic analysis within the field of male reproduction.


\bibliography{bibliography.bib}

\begin{thebibliography}{10}
\urlstyle{rm}
\expandafter\ifx\csname url\endcsname\relax
  \def\url#1{\texttt{#1}}\fi
\expandafter\ifx\csname urlprefix\endcsname\relax\def\urlprefix{URL }\fi
\expandafter\ifx\csname doiprefix\endcsname\relax\def\doiprefix{DOI: }\fi
\providecommand{\bibinfo}[2]{#2}
\providecommand{\eprint}[2][]{\url{#2}}

\bibitem{topol2019high}
\bibinfo{author}{Topol, E.~J.}
\newblock \bibinfo{journal}{\bibinfo{title}{High-performance medicine: the
  convergence of human and artificial intelligence}}.
\newblock {\emph{\JournalTitle{Nature medicine}}}
  \textbf{\bibinfo{volume}{25}}, \bibinfo{pages}{44} (\bibinfo{year}{2019}).

\bibitem{boll2018health}
\bibinfo{author}{Boll, S.}, \bibinfo{author}{Meyer, J.} \&
  \bibinfo{author}{O’Connor, N.~E.}
\newblock \bibinfo{journal}{\bibinfo{title}{Health media: From multimedia
  signals to personal health insights}}.
\newblock {\emph{\JournalTitle{IEEE MultiMedia}}}
  \textbf{\bibinfo{volume}{25}}, \bibinfo{pages}{51--60}
  (\bibinfo{year}{2018}).

\bibitem{Riegler2016}
\bibinfo{author}{Riegler, M.} \emph{et~al.}
\newblock \bibinfo{title}{Multimedia and medicine: Teammates for better disease
  detection and survival}.
\newblock In \emph{\bibinfo{booktitle}{Proceedings of the ACM International
  Conference on Multimedia (ACM MM)}}, \bibinfo{pages}{968--977},
  \doiprefix\url{10.1145/2964284.2976760} (\bibinfo{publisher}{ACM},
  \bibinfo{year}{2016}).

\bibitem{Hannun2019}
\bibinfo{author}{Hannun, A.~Y.} \emph{et~al.}
\newblock \bibinfo{journal}{\bibinfo{title}{{Cardiologist-level arrhythmia
  detection and classification in ambulatory electrocardiograms using a deep
  neural network}}}.
\newblock {\emph{\JournalTitle{Nature Medicine}}}
  \textbf{\bibinfo{volume}{25}}, \bibinfo{pages}{65--69},
  \doiprefix\url{10.1038/s41591-018-0268-3} (\bibinfo{year}{2019}).

\bibitem{esteva2017}
\bibinfo{author}{Esteva, A.} \emph{et~al.}
\newblock \bibinfo{journal}{\bibinfo{title}{{Dermatologist-level classification
  of skin cancer with deep neural networks}}}.
\newblock {\emph{\JournalTitle{Nature}}} \textbf{\bibinfo{volume}{542}},
  \bibinfo{pages}{115--118}, \doiprefix\url{10.1038/nature21056}
  (\bibinfo{year}{2017}).

\bibitem{Pogorelov2017}
\bibinfo{author}{Pogorelov, K.} \emph{et~al.}
\newblock \bibinfo{journal}{\bibinfo{title}{Efficient disease detection in
  gastrointestinal videos -- global features versus neural networks}}.
\newblock {\emph{\JournalTitle{Multimedia Tools and Applications}}}
  \textbf{\bibinfo{volume}{76}}, \bibinfo{pages}{22493--22525},
  \doiprefix\url{10.1007/s11042-017-4989-y} (\bibinfo{year}{2017}).

\bibitem{carlsen1992evidence}
\bibinfo{author}{Carlsen, E.}, \bibinfo{author}{Giwercman, A.},
  \bibinfo{author}{Keiding, N.} \& \bibinfo{author}{Skakkeb{\ae}k, N.~E.}
\newblock \bibinfo{journal}{\bibinfo{title}{Evidence for decreasing quality of
  semen during past 50 years.}}
\newblock {\emph{\JournalTitle{British Medical Journal}}}
  \textbf{\bibinfo{volume}{305}}, \bibinfo{pages}{609--613}
  (\bibinfo{year}{1992}).

\bibitem{levine2017temporal}
\bibinfo{author}{Levine, H.} \emph{et~al.}
\newblock \bibinfo{journal}{\bibinfo{title}{Temporal trends in sperm count: a
  systematic review and meta-regression analysis}}.
\newblock {\emph{\JournalTitle{Human Reproduction Update}}}
  \textbf{\bibinfo{volume}{23}}, \bibinfo{pages}{646--659}
  (\bibinfo{year}{2017}).

\bibitem{jorgensen2002east}
\bibinfo{author}{J{\o}rgensen, N.} \emph{et~al.}
\newblock \bibinfo{journal}{\bibinfo{title}{East--west gradient in semen
  quality in the nordic--baltic area: a study of men from the general
  population in denmark, norway, estonia and finland}}.
\newblock {\emph{\JournalTitle{Human Reproduction}}}
  \textbf{\bibinfo{volume}{17}}, \bibinfo{pages}{2199--2208}
  (\bibinfo{year}{2002}).

\bibitem{tomlinson2016uncertainty}
\bibinfo{author}{Tomlinson, M.}
\newblock \bibinfo{journal}{\bibinfo{title}{Uncertainty of measurement and
  clinical value of semen analysis: has standardisation through professional
  guidelines helped or hindered progress?}}
\newblock {\emph{\JournalTitle{Andrology}}} \textbf{\bibinfo{volume}{4}},
  \bibinfo{pages}{763--770} (\bibinfo{year}{2016}).

\bibitem{WHOmanual2010laboratory}
\bibinfo{author}{{World Health Organization, Department of Reproductive Health
  and Research}}.
\newblock \emph{\bibinfo{title}{WHO laboratory manual for the examination and
  processing of human semen}} (\bibinfo{publisher}{Geneva: World Health
  Organization}, \bibinfo{year}{2010}).

\bibitem{cooper2009world}
\bibinfo{author}{Cooper, T.~G.} \emph{et~al.}
\newblock \bibinfo{journal}{\bibinfo{title}{World health organization reference
  values for human semen characteristics}}.
\newblock {\emph{\JournalTitle{Human Reproduction Update}}}
  \textbf{\bibinfo{volume}{16}}, \bibinfo{pages}{231--245},
  \doiprefix\url{10.1093/humupd/dmp048} (\bibinfo{year}{2010}).

\bibitem{mortimer2015future}
\bibinfo{author}{Mortimer, S.~T.}, \bibinfo{author}{van~der Horst, G.} \&
  \bibinfo{author}{Mortimer, D.}
\newblock \bibinfo{journal}{\bibinfo{title}{The future of computer-aided sperm
  analysis}}.
\newblock {\emph{\JournalTitle{Asian journal of andrology}}}
  \textbf{\bibinfo{volume}{17}}, \bibinfo{pages}{545} (\bibinfo{year}{2015}).

\bibitem{dearing2014validation}
\bibinfo{author}{Dearing, C.~G.}, \bibinfo{author}{Kilburn, S.} \&
  \bibinfo{author}{Lindsay, K.~S.}
\newblock \bibinfo{journal}{\bibinfo{title}{Validation of the sperm class
  analyser casa system for sperm counting in a busy diagnostic semen analysis
  laboratory}}.
\newblock {\emph{\JournalTitle{Human Fertility}}}
  \textbf{\bibinfo{volume}{17}}, \bibinfo{pages}{37--44}
  (\bibinfo{year}{2014}).

\bibitem{dearing2019can}
\bibinfo{author}{Dearing, C.}, \bibinfo{author}{Jayasena, C.} \&
  \bibinfo{author}{Lindsay, K.}
\newblock \bibinfo{journal}{\bibinfo{title}{Can the sperm class analyser (sca)
  casa-mot system for human sperm motility analysis reduce imprecision and
  operator subjectivity and improve semen analysis?}}
\newblock {\emph{\JournalTitle{Human Fertility}}} \bibinfo{pages}{1--11}
  (\bibinfo{year}{2019}).

\bibitem{Urbano2017}
\bibinfo{author}{{Urbano}, L.~F.}, \bibinfo{author}{{Masson}, P.},
  \bibinfo{author}{{VerMilyea}, M.} \& \bibinfo{author}{{Kam}, M.}
\newblock \bibinfo{journal}{\bibinfo{title}{Automatic tracking and motility
  analysis of human sperm in time-lapse images}}.
\newblock {\emph{\JournalTitle{IEEE Transactions on Medical Imaging}}}
  \textbf{\bibinfo{volume}{36}}, \bibinfo{pages}{792--801},
  \doiprefix\url{10.1109/TMI.2016.2630720} (\bibinfo{year}{2017}).

\bibitem{Dewan2018}
\bibinfo{author}{Dewan, K.}, \bibinfo{author}{Rai~Dastidar, T.} \&
  \bibinfo{author}{Ahmad, M.}
\newblock \bibinfo{title}{Estimation of sperm concentration and total motility
  from microscopic videos of human semen samples}.
\newblock In \emph{\bibinfo{booktitle}{Proceedings of the IEEE Conference on
  Computer Vision and Pattern Recognition (CVPR) Workshops}}
  (\bibinfo{year}{2018}).

\bibitem{ghasemian2015efficient}
\bibinfo{author}{Ghasemian, F.}, \bibinfo{author}{Mirroshandel, S.~A.},
  \bibinfo{author}{Monji-Azad, S.}, \bibinfo{author}{Azarnia, M.} \&
  \bibinfo{author}{Zahiri, Z.}
\newblock \bibinfo{journal}{\bibinfo{title}{An efficient method for automatic
  morphological abnormality detection from human sperm images}}.
\newblock {\emph{\JournalTitle{Computer methods and programs in biomedicine}}}
  \textbf{\bibinfo{volume}{122}}, \bibinfo{pages}{409--420}
  (\bibinfo{year}{2015}).

\bibitem{shaker2017dictionary}
\bibinfo{author}{Shaker, F.}, \bibinfo{author}{Monadjemi, S.~A.},
  \bibinfo{author}{Alirezaie, J.} \& \bibinfo{author}{Naghsh-Nilchi, A.~R.}
\newblock \bibinfo{journal}{\bibinfo{title}{A dictionary learning approach for
  human sperm heads classification}}.
\newblock {\emph{\JournalTitle{Computers in biology and medicine}}}
  \textbf{\bibinfo{volume}{91}}, \bibinfo{pages}{181--190}
  (\bibinfo{year}{2017}).

\bibitem{visem}
\bibinfo{author}{Haugen, T.} \emph{et~al.}
\newblock \bibinfo{title}{Visem: A multimodal video dataset of human
  spermatozoa}.
\newblock In \emph{\bibinfo{booktitle}{Proceedings of the ACM Multimedia
  Systems Conference (MMSYS)}}, \doiprefix\url{10.1145/3304109.3325814}
  (\bibinfo{publisher}{ACM}, \bibinfo{year}{2019}).

\bibitem{andersen2015body}
\bibinfo{author}{Andersen, J.~M.} \emph{et~al.}
\newblock \bibinfo{journal}{\bibinfo{title}{Body mass index is associated with
  impaired semen characteristics and reduced levels of anti-m{\"u}llerian
  hormone across a wide weight range}}.
\newblock {\emph{\JournalTitle{PloS one}}} \textbf{\bibinfo{volume}{10}},
  \bibinfo{pages}{e0130210} (\bibinfo{year}{2015}).

\bibitem{nadeau2000inference}
\bibinfo{author}{Nadeau, C.} \& \bibinfo{author}{Bengio, Y.}
\newblock \bibinfo{title}{Inference for the generalization error}.
\newblock In \emph{\bibinfo{booktitle}{Proceeding of the Advances in neural
  information processing systems (NIPS)}}, \bibinfo{pages}{307--313}
  (\bibinfo{year}{2000}).

\bibitem{lux2016lire}
\bibinfo{author}{Lux, M.}, \bibinfo{author}{Riegler, M.},
  \bibinfo{author}{Halvorsen, P.}, \bibinfo{author}{Pogorelov, K.} \&
  \bibinfo{author}{Anagnostopoulos, N.}
\newblock \bibinfo{title}{Lire: open source visual information retrieval}.
\newblock In \emph{\bibinfo{booktitle}{Proceedings of the ACM Multimedia
  Systems Conference (MMSYS)}}, \bibinfo{pages}{30} (\bibinfo{year}{2016}).

\bibitem{Weka2009}
\bibinfo{author}{Hall, M.} \emph{et~al.}
\newblock \bibinfo{journal}{\bibinfo{title}{The {WEKA} data mining software: an
  update}}.
\newblock {\emph{\JournalTitle{SIGKDD Explorations}}}
  \textbf{\bibinfo{volume}{11}}, \bibinfo{pages}{10--18}
  (\bibinfo{year}{2009}).

\bibitem{Dozat2015IncorporatingNM}
\bibinfo{author}{Dozat, T.}
\newblock \bibinfo{title}{Incorporating nesterov momentum into adam}
  (\bibinfo{year}{2015}).

\bibitem{huang2017densely}
\bibinfo{author}{Huang, G.}, \bibinfo{author}{Liu, Z.},
  \bibinfo{author}{van~der Maaten, L.} \& \bibinfo{author}{Weinberger, K.~Q.}
\newblock \bibinfo{title}{Densely connected convolutional networks}.
\newblock In \emph{\bibinfo{booktitle}{Proceedings of the IEEE Conference on
  Computer Vision and Pattern Recognition (CVPR)}} (\bibinfo{year}{2017}).

\bibitem{He2016DeepRL}
\bibinfo{author}{He, K.}, \bibinfo{author}{Zhang, X.}, \bibinfo{author}{Ren,
  S.} \& \bibinfo{author}{Sun, J.}
\newblock \bibinfo{title}{Deep residual learning for image recognition}.
\newblock \bibinfo{pages}{770--778} (\bibinfo{year}{2016}).

\bibitem{szegedy2015rethinking}
\bibinfo{author}{Szegedy, C.}, \bibinfo{author}{Vanhoucke, V.},
  \bibinfo{author}{Ioffe, S.}, \bibinfo{author}{Shlens, J.} \&
  \bibinfo{author}{Wojna, Z.}
\newblock \bibinfo{journal}{\bibinfo{title}{Rethinking the inception
  architecture for computer vision}}.
\newblock {\emph{\JournalTitle{arXiv preprint arXiv:1512.00567}}}
  (\bibinfo{year}{2015}).

\bibitem{imagenetcvpr09}
\bibinfo{author}{Deng, J.} \emph{et~al.}
\newblock \bibinfo{title}{{ImageNet: A Large-Scale Hierarchical Image
  Database}}.
\newblock In \emph{\bibinfo{booktitle}{Proceedings of the IEEE Conference on
  Computer Vision and Pattern Recognition (CVPR)}} (\bibinfo{year}{2009}).

\bibitem{chollet2015keras}
\bibinfo{author}{Chollet, F.} \emph{et~al.}
\newblock \bibinfo{title}{Keras: Deep learning library for theano and
  tensorflow}.
\newblock \bibinfo{howpublished}{\url{https://keras.io}}
  (\bibinfo{year}{2015}).

\bibitem{Tensorflow2015}
\bibinfo{author}{Abadi, M.} \emph{et~al.}
\newblock \bibinfo{title}{Tensorflow: A system for large-scale machine
  learning}.
\newblock In \emph{\bibinfo{booktitle}{12th USENIX Symposium on Operating
  Systems Design and Implementation (OSDI 16)}}, \bibinfo{pages}{265--283}
  (\bibinfo{year}{2016}).

\bibitem{Lucas1981}
\bibinfo{author}{Lucas, B.~D.} \& \bibinfo{author}{Kanade, T.}
\newblock \bibinfo{title}{An iterative image registration technique with an
  application to stereo vision}.
\newblock In \emph{\bibinfo{booktitle}{Proceedings of the International Joint
  Conference on Artificial Intelligence (IJCAI) - Volume 2}},
  \bibinfo{pages}{674--679} (\bibinfo{publisher}{Morgan Kaufmann Publishers
  Inc.}, \bibinfo{year}{1981}).

\bibitem{Harris88acombined}
\bibinfo{author}{Harris, C.} \& \bibinfo{author}{Stephens, M.}
\newblock \bibinfo{title}{A combined corner and edge detector}.
\newblock In \emph{\bibinfo{booktitle}{Proceedings of the Alvey Vision
  Conference}}, \bibinfo{pages}{147--151} (\bibinfo{year}{1988}).

\bibitem{opencv_library}
\bibinfo{author}{Bradski, G.}
\newblock \bibinfo{journal}{\bibinfo{title}{{The OpenCV Library}}}.
\newblock {\emph{\JournalTitle{Dr. Dobb's Journal of Software Tools}}}
  (\bibinfo{year}{2000}).

\bibitem{Gunnar2003}
\bibinfo{author}{Farneb{\"a}ck, G.}
\newblock \bibinfo{title}{Two-frame motion estimation based on polynomial
  expansion}.
\newblock In \bibinfo{editor}{Bigun, J.} \& \bibinfo{editor}{Gustavsson, T.}
  (eds.) \emph{\bibinfo{booktitle}{Image Analysis}}, \bibinfo{pages}{363--370}
  (\bibinfo{publisher}{Springer Berlin Heidelberg}, \bibinfo{address}{Berlin,
  Heidelberg}, \bibinfo{year}{2003}).

\bibitem{simonyan2014two}
\bibinfo{author}{Simonyan, K.} \& \bibinfo{author}{Zisserman, A.}
\newblock \bibinfo{title}{Two-stream convolutional networks for action
  recognition in videos}.
\newblock In \emph{\bibinfo{booktitle}{Proceedings of Advances in Neural
  Information Processing Systems (NIPS)}}, \bibinfo{pages}{568--576}
  (\bibinfo{year}{2014}).

\end{thebibliography}




\section*{Author Contributions Statement}
S.A.H, M.A.R and T.B.H conceived the experiment(s),  S.A.H., V.T. and M.A.R. conducted the experiment(s), all authors analysed the results. All authors reviewed the manuscript. 

\section*{Data Availability Statement}
The dataset used for all experiments is publicly available at https://datasets.simula.no/visem/ for non-commercial use. The data is fully anonymized (no keys for re-identification are stored).

\section*{Ethical approval and Informed Consent}
In this study, we used fully anonymized data originally collected based on written informed consent and approval by the Regional Committee for Medical and Health Research Ethics - South East Norway. Furthermore, we confirm that all experiments were performed in accordance with the relevant guidelines and regulations of the Regional Committee for Medical and Health Research Ethics - South East Norway, and the \gls{gdpr}.

\section*{Additional information}
\textbf{Competing interests} The author(s) declare no competing interests.

\end{document}